# Development of Multistage Machine Learning Classifier using Decision Trees and Boosting Algorithms over Darknet Network Traffic

Anjali Sureshkumar Nair
Department of Information Technology
Pillai College of Engineering,
University of Mumbai
anjali22meit@student.mes.ac.in

Dr. Prashant Nitnaware
Department of Computer Engineering
Pillai College of Engineering,
University of Mumbai
pnitnaware@mes.ac.in

*Abstract—* **In recent years, the clandestine nature of darknet activities has presented an escalating challenge to cybersecurity efforts, necessitating sophisticated methods for the detection and classification of network traffic associated with these covert operations. The system addresses the significant challenge of class imbalance within Darknet traffic datasets, where malicious traffic constitutes a minority, hindering effective discrimination between normal and malicious behavior. By leveraging boosting algorithms like AdaBoost and Gradient Boosting coupled with decision trees, this study proposes a robust solution for network traffic classification. Boosting algorithms ensemble learning corrects errors iteratively and assigns higher weights to minority class instances, complemented by the hierarchical structure of decision trees. The additional Feature Selection which is a pre-processing method by utilizing Information Gain metrics, Fisher's Score, and Chi-Square test selection for features is employed. Rigorous experimentation with diverse Darknet traffic datasets validates the efficacy of the proposed multistage classifier, evaluated through various performance metrics such as accuracy, precision, recall, and F1-score, offering a comprehensive solution for accurate detection and classification of Darknet activities.**

I. INTRODUCTION

The escalating complexity and diversity of darknet network activities demand innovative approaches for effective identification and classification. Leveraging the synergies between decision trees and boosting algorithms, our proposed classifier exhibits a robust capability to discern and categorize intricate patterns inherent in darknet activities.

The proliferation of darknet activities in the digital realm has cast a shadow over the cybersecurity landscape, necessitating the development of advanced techniques for identifying and classifying network traffic associated with these clandestine operations. The darknet, characterized by its anonymity-centric infrastructure, serves as a breeding ground for various illicit activities, including but not limited to cybercrime, illegal trade, and information warfare. As a result, the imperative to fortify cybersecurity measures against these elusive threats has become paramount.

The development of this multistage machine learning classifier represents a significant advancement in the field of darknet network traffic analysis. The integration of decision trees and boosting algorithms, coupled with meticulous feature engineering, positions the classifier as a robust solution for addressing the challenges posed by the clandestine nature of darknet activities. This research contributes to the ongoing efforts to bolster cybersecurity measures and underscores the importance of innovative machine-learning approaches in safeguarding network integrity in the face of evolving threats.

The primary objective of this research is to advance the state-of-the-art in darknet network traffic analysis by introducing a multistage machine learning classifier that combines decision trees and boosting algorithms. This hybrid approach aims to improve the accuracy, precision, and adaptability of classification models, thereby enhancing the capacity to detect and classify a diverse range of darknet activities. The research also seeks to address the limitations of existing single-stage models, which may struggle to capture the complexity and variability inherent in darknet network traffic patterns.

The overarching objective is to confront the growing challenges posed by darknet network traffic and the increasingly sophisticated tactics employed by malicious actors within these covert online spaces. Motivated by the imperative to enhance network security amidst evolving cyber threats, the study focuses on evaluating the efficiency of machine learning algorithms in multilayered classification to bolster threat detection capabilities. Key motivating factors include the emergence of darknet as a breeding ground for illicit activities, the perpetual evolution of malicious tactics necessitating advanced detection tools, and the critical need for accurate threat detection to enable proactive cybersecurity measures. Additionally, the intricate nature of darknet network traffic underscores the importance of adopting a multilayered classification approach, highlighting the significance of exploring the efficacy of machine learning algorithms in addressing these challenges and crafting comprehensive detection systems.

With a comprehensive set of objectives, the project aims to rigorously evaluate, design, optimize, and validate machine learning algorithms and multilayered classification techniques tailored for darknet network traffic. By achieving these objectives, the study seeks to contribute significantly to the development of more precise and efficient systems capable of discerning and categorizing malicious activities within darknet environments. Specifically, the research focuses on assessing the performance and effectiveness of Decision Tree and Boosting algorithms in accurately classifying various types of darknet network traffic, utilizing metrics such as accuracy, precision, recall, and F1 score to measure classification effectiveness. Furthermore, the study may delve into exploring the influence of different features, feature selection methods, and preprocessing techniques on algorithm performance, providing valuable insights into optimizing the classification process for multilayered darknet traffic.

In essence, the proposed system represents a concerted effort to confront the complex challenges posed by darknet network traffic through the lens of machine learning. By leveraging advanced algorithms and multilayered classification approaches, the study endeavors to advance threat detection capabilities, fortify network security measures and contribute to the ongoing fight against cyber threats in an ever-evolving digital landscape.

## II. LITERATURE SURVEY

Several recent studies have investigated the effectiveness of multiclass ML classifiers in various domains. (Dey and Pratap, 2023) conducted a comparative study of oversampling techniques such as SMOTE, Borderline SMOTE, and ADASYN, highlighting their impact on classification accuracy across different datasets. However, their study primarily focused on oversampling techniques and lacked exploration into alternative classification algorithms and also oversampling can at times lead to overfitting issues. (Yang et al., 2023) proposed a novel multi-feature fusion approach for classifying encrypted mobile application traffic, achieving high accuracy rates. Nonetheless, their study did not thoroughly address feature fusion's limitations or potential biases.

(Karagol et al., 2022) introduced ML-based traffic classification approaches using SMOTE for class balancing and feature selection algorithms. While their methods showed promising accuracy rates, they did not extensively explore the scalability or generalization of their models to different traffic datasets. (Wang et al., 2022) presented a differential preserving approach in XGBoost model for encrypted traffic classification, maintaining data privacy without decryption. However, their study did not discuss the potential computational overhead or performance degradation associated with preserving data differentials.

(Chen et al., (2022) utilized generative models to classify encapsulated and anonymized network video traffic, achieving high accuracy rates. Nonetheless, their study did not thoroughly investigate the robustness of their models to adversarial attacks or network anomalies. (Xu et al., 2022) proposed an encrypted traffic classification method based on path signature features, demonstrating superior performance compared to existing methods. However, their study did not address the potential limitations or biases introduced by the selection of path transformations.

(Iliadis and Kaifas, 2021) employed common ML classification algorithms for darknet traffic classification, achieving high prediction accuracy. Yet, their study lacked exploration into the scalability or adaptability of their models to evolving network threats. (Gupta, 2021) introduced a deep reinforcement learning approach for VPN-non-VPN traffic classification, achieving competitive accuracy rates. However, their study did not thoroughly investigate the generalization of their models to different network environments or traffic patterns.

(Cai et al., 2021) proposed a Mobile Encrypted Traffic Classification method using Markov Chains and Graph Neural Networks, achieving improved accuracy and reduced training time. However, their study did not address the potential biases introduced by the selection of graph representations or the scalability of their models to large-scale network traffic.

(Aswad and Sonuç, 2020) utilized artificial neural networks and Apache Spark for VPN traffic classification, achieving high precision rates. However, their study did not thoroughly investigate the scalability or real-time performance of their models in large-scale network environments. (Baldini, 2020) applied time-frequency analysis to encrypted traffic classification, revealing insights into traffic patterns. However, their study did not thoroughly investigate the privacy implications or potential vulnerabilities associated with feature analysis.

Overall, while these studies have made significant contributions to the field of multiclass ML classifiers for traffic classification, there remains a need for further research to address the identified drawbacks and limitations, including biases in feature selection, scalability issues, and privacy concerns.

## III. RELATED WORKS

Previous studies have utilized decision trees for traffic classification, such as (Karagol et al., 2022). However, these studies often lack comprehensive feature selection methods, leading to potential biases and overfitting. Boosting algorithms, as demonstrated by (Wang et al., 2022), have shown promise in accurately classifying encrypted traffic. Nonetheless, there is a need to address potential computational overhead and performance degradation associated with preserving data differentials.

In terms of feature selection techniques, (Yang et al., 2023) proposed a multi-feature fusion approach. However, the study did not thoroughly address limitations or biases associated with feature fusion. Therefore, we can explore more robust feature selection methods, such as the Relief-F algorithm or differential preserving, to enhance model performance and interpretability. The proposed system aims to classify darknet traffic using decision trees and boosting algorithms while addressing the identified drawbacks in the existing systems.

The existing systems highlighted in the literature survey exhibit notable advancements in the classification of encrypted and darknet network traffic. However, several drawbacks and limitations are apparent, hindering their efficacy in certain scenarios. For instance, while some systems demonstrate high accuracy in classifying specific types of traffic, they may lack scalability or fail to generalize well to diverse datasets. Additionally, certain systems rely heavily on complex machine learning algorithms, leading to increased computational overhead and resource requirements, which may not be feasible for real-time deployment in large-scale networks.

Moreover, some systems face challenges in feature selection and extraction, potentially overlooking crucial characteristics of encrypted traffic patterns. This limitation can impact the system's ability to differentiate between benign and malicious activities accurately. Furthermore, the lack of interpretability in certain models poses challenges for cybersecurity analysts in understanding and validating classification results, thereby limiting their trust and adoption. Furthermore, existing systems may struggle with the dynamic nature of network traffic and evolving threat landscapes. They may not effectively adapt to changing traffic patterns or detect novel attack vectors, compromising their ability to provide timely and accurate threat intelligence. Additionally, the reliance on static datasets for training and evaluation may limit the system's ability to generalize to real-world scenarios with diverse and evolving network environments.

Overall, while existing systems demonstrate promising capabilities in encrypted and darknet traffic classification, they are not without limitations. Addressing these drawbacks requires a holistic approach that considers scalability, interpretability, adaptability, and generalization capabilities to ensure robust and effective cybersecurity solutions in dynamic network environments.

## IV. PROPOSED SYSTEM

FIGURE I. MULTISTAGE DARKNET TRAFFIC CLASSIFIER (DTC) SYSTEM

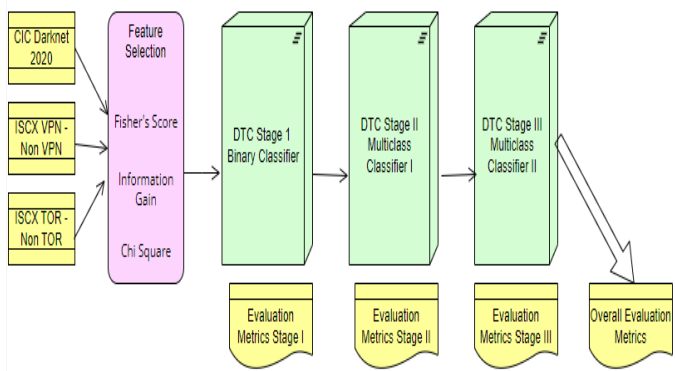

Fig. 1. Multistage DTC Classifier System.

a) **Datasets:** Input datasets consist of three datasets namely CIC-Darknet2020, ISCX-VPN2016, and ISCX-Tor2016.

b) **Feature Selection:** Feature Selection algorithms like Fisher's Score (to rank the features as per their continuous feature score), Information Gain (to compute discriminant ratio among features), and Chi-Square Test algorithms (to find the association among categorical features) are utilized for Feature Selection.

c) **Multistage Darknet Traffic Classifier (DTC) system:**

1. DTC Stage I (Binary Classifier) – Binary Classifier classifies Darknet network traffic into Benign and Malicious by utilizing AdaBoost, KNN (K-Nearest Neighbors), and Decision Trees algorithms.

2. DTC Stage II (Multiclass Classifier I) – Multiclass I Classifier classifies Darknet network traffic into Tor (The Onion Router), Non-Tor, VPN (Virtual Private Network), and Non-VPN by utilizing AdaBoost, Random Forest, Naïve Bayes, KNN (K-Nearest Neighbors), Decision Trees and Gradient Boosting algorithms.

3. DTC Stage III (Multiclass Classifier II) – Multiclass II Classifier classifies Darknet network traffic into eight application classes – File transfer, Audio stream, P2P (Peer-to-peer services), Browsing, Video stream, Chat, Email, and VoIP (Voice over Internet Protocol) by utilizing AdaBoost, Random Forest, and Decision Trees algorithms.

Firstly, we will gather darknet traffic data from reputable sources or datasets like CIC-Darknet2020, ISCX-VPN2016, ISCX-Tor2016 and preprocess the data to handle missing values, normalize features, and extract relevant traffic attributes. Next, we will employ advanced feature selection techniques, including Fisher's Score, Information Gain, and Chi-Square Test algorithms, to identify discriminative features. For model development, we will utilize decision tree algorithms like decision trees, Random Forests, and boosting algorithms such as Gradient Boosting and AdaBoost to improve classification accuracy and handle imbalanced datasets.

In terms of evaluation and validation, we will assess model performance using metrics like accuracy, recall, F1 score, and precision. We will also conduct cross-validation and performance comparisons with existing approaches to validate the effectiveness of the proposed system. To address drawbacks, we will mitigate biases introduced by feature selection through robust validation methods and interpretability analysis. Additionally, we will address potential computational overhead by optimizing model training and inference processes. Furthermore, we will compare the results after each stage along with the overall result after 3 stages of Multistage Darknet Traffic Classifier (DTC Stage III – Multiclass Classifier II). By incorporating these components into our proposed system, we aim to develop a robust and scalable solution for darknet traffic classification while addressing the identified drawbacks in existing literature.

Incorporating decision trees and boosting algorithms instead of complex hybrid systems for darknet, VPN, and Tor classifications offers several advantages in terms of effectiveness and usefulness. Decision trees provide a transparent and easy-to-understand decision-making process, making them highly interpretable. This is beneficial for security analysts who need to understand how classifications are made and interpret the results effectively. Additionally, decision trees have relatively low computational complexity compared to more complex algorithms, making them efficient for real-time or high-throughput classification tasks. This efficiency is crucial for handling large volumes of network traffic data promptly.

Boosting algorithms, such as XGBoost or AdaBoost, enhance the performance of decision trees by iteratively correcting the errors of weak learners. This allows them to capture complex non-linear relationships in the data without the need for complex hybrid systems. Moreover, decision trees and boosting algorithms are robust against noise and irrelevant features in the data. They can handle mixed types of data, including categorical and numerical variables, without requiring extensive preprocessing or feature engineering. This robustness ensures reliable classification results even in noisy or unstructured datasets.

Furthermore, decision trees and boosting algorithms are scalable and can handle large datasets with millions of samples and thousands of features. This scalability makes them suitable for processing network traffic data collected from diverse sources and environments. Additionally, these algorithms offer flexibility in adapting to different classification tasks and objectives. Security analysts can easily customize the classification models according to specific requirements and preferences. Overall, by incorporating decision trees and boosting algorithms for darknet, VPN, and Tor classifications, organizations can achieve effective and efficient network traffic analysis while maintaining simplicity, interpretability, and scalability. These algorithms offer a balance between performance and complexity, making them valuable tools for cybersecurity applications.

The system adeptly addresses a myriad of applicative fields, showcasing its comprehensive approach to bolstering cybersecurity measures. Leveraging decision trees and boosting algorithms, the system seamlessly responds to diverse challenges across various domains. In threat detection and intrusion prevention, it accurately classifies darknet network traffic, swiftly identifying malicious activities such as botnet operations, malware distribution, and illegal marketplaces. For anomaly detection, the system effectively discerns deviations from normal behavior, facilitating prompt responses to potential security breaches. In cybercrime investigation, it provides crucial insights into darknet activities, aiding law enforcement agencies in combating criminal operations effectively by deriving overall accuracy and better classification performance in classifying datasets that belong to the Darknet or Dark web nature. Furthermore, the system contributes to risk assessment, threat intelligence, network monitoring, SIEM (Security Information and Event Management) enhancement, malware analysis, and proactive defense measures. Its robust classification capabilities empower organizations to strengthen their cybersecurity posture and proactively defend against emerging threats from the darknet across a spectrum of applications.

## V. PERFORMANCE COMPARISON OF DECISION TREES, BOOSTING ALGORITHMS STAGE-WISE

TABLE I. COMPARISON OF EVALUATION METRICS STAGE-WISE

| Algorithms | Comparison of Evaluation Metrics Stage-wise | | | | |
|---|---|---|---|---|---|
| | DTC Stages | Accuracy | F1 Score | Precision | Recall |
| AdaBoost | DTC I | 99.7491 | 99.5543 | 99.3073 | 99.8056 |
| KNN | DTC I | 99.6467 | 99.4157 | 99.3754 | 99.4561 |
| Decision Trees | DTC I | 99.9594 | 98.8655 | 98.1911 | 99.5729 |
| AdaBoost | DTC2 | 88.0626 | 68.3520 | 70.1264 | 70.9164 |
| Random Forest | DTC2 | 99.9964 | 99.9534 | 99.9122 | 99.9947 |
| Naive Bayes | DTC2 | 85.9711 | 68.4271 | 71.0140 | 80.3620 |
| Gradient Boosting | DTC2 | 99.6007 | 98.2755 | 97.9536 | 98.6066 |
| Random Forest | DTC3 | 99.6113 | 97.7945 | 99.9122 | 99.9947 |
| AdaBoost | DTC3 | 99.6855 | 97.5292 | 99.7272 | 99.8949 |
| KNN | DTC2 | 99.6184 | 98.5490 | 98.3574 | 98.7433 |
| Decision Trees | DTC2 | 99.9823 | 99.8107 | 99.7272 | 99.8949 |
| Decision Trees | DTC3 | 99.4453 | 97.1733 | 98.8455 | 96.0276 |

Fig. 2. Performance comparison of Machine Learning algorithms over Multiclassifier DTC (Darknet Traffic Classifier) Stages for Darknet 2020 dataset.

Using decision trees and boosting algorithms offers several benefits over other machine learning algorithms in terms of various evaluation metrics:

**Accuracy:** Decision trees and boosting algorithms, such as AdaBoost and Gradient Boosting, often achieve high accuracy in classification tasks. They can effectively partition the feature space and make decisions based on simple rules, leading to accurate predictions.

**Precision and Recall:** Decision trees and boosting algorithms tend to have good precision and recall rates. They can effectively identify positive instances (precision) while minimizing false negatives (recall), making them suitable for tasks where class imbalance is a concern.

**F1 Score:** Decision trees and boosting algorithms typically yield high F1 scores, which provide a balance between precision and recall. This metric is particularly useful in binary classification tasks and is indicative of the model's overall performance.

**Robustness to Noise:** Decision trees and boosting algorithms are robust to noisy data and outliers. They can handle nonlinear relationships between features and class labels, making them suitable for complex datasets with non-trivial decision boundaries.

**Interpretability:** Decision trees offer interpretability, as the decision-making process is transparent and can be easily visualized. Boosting algorithms, although more complex, can still provide insights into feature importance and model behavior, aiding in model interpretation.

**Ensemble Learning:** Boosting algorithms, such as AdaBoost and Gradient Boosting, are ensemble methods that combine multiple weak learners to create a strong classifier. This ensemble approach often leads to improved generalization performance and reduced overfitting.

**Scalability:** Decision trees and boosting algorithms are generally scalable and can handle large datasets efficiently. They can be parallelized and distributed across multiple processors or nodes, allowing for faster training and inference times.

Overall, decision trees and boosting algorithms offer a powerful framework for classification tasks, providing high accuracy, robustness, interpretability, and scalability, making them well-suited for a wide range of applications in cybersecurity, finance, healthcare, and more.

## VI. EXPERIMENTAL RESULTS

| Algorithms | DTC (Darknet Traffic Classifier) Stages | Accuracy | F1 Score | Precision | Recall |
|---|---|---|---|---|---|
| AdaBoost | DTC 1 | 99.7491 | 99.5558 | 99.3073 | 99.8056 |
| KNN | DTC 1 | 99.6467 | 99.4157 | 99.3754 | 99.4561 |
| Decision Trees | DTC 1 | 99.9594 | 98.8772 | 98.1911 | 99.5729 |
| AdaBoost | DTC2 | 88.0626 | 70.5192 | 70.1264 | 70.9164 |
| Random Forest | DTC2 | 99.9964 | 99.9534 | 99.9122 | 99.9947 |
| Naive Bayes | DTC2 | 85.9711 | 75.3994 | 71.0140 | 80.3620 |
| Gradient Boosting | DTC2 | 99.6007 | 98.2790 | 97.9536 | 98.6066 |
| KNN | DTC2 | 99.6184 | 98.5500 | 98.3574 | 98.7433 |
| Decision Trees | DTC2 | 99.9823 | 99.8110 | 99.7272 | 99.8949 |
| Decision Trees | DTC3 | 99.4453 | 97.4162 | 98.8455 | 96.0276 |
| Random Forest | DTC3 | 99.6113 | 99.9534 | 99.9122 | 99.9947 |
| AdaBoost | DTC3 | 99.6855 | 99.8110 | 99.7272 | 99.8949 |

Fig. 3. Comparison of Evaluation metrics over input dataset CIC-Darknet2020 in Multiclassifier Darknet Traffic Classifier (DTC) Stages.

| Algorithms | DTC (Darknet Traffic Classifier) Stages | Accuracy | F1 Score | Precision | Recall |
|---|---|---|---|---|---|
| AdaBoost | DTC 1 | 99.7067 | 99.4879 | 99.2197 | 99.7576 |
| KNN | DTC 1 | 99.7209 | 99.5085 | 99.4803 | 99.5369 |
| Decision Trees | DTC 1 | 99.8266 | 98.8814 | 98.2303 | 99.5411 |
| AdaBoost | DTC2 | 82.2758 | 71.3916 | 68.9776 | 73.9809 |
| Random Forest | DTC2 | 99.9964 | 99.9548 | 99.9947 | 99.9149 |
| Naive Bayes | DTC2 | 86.3951 | 75.9897 | 71.7448 | 80.7685 |
| Gradient Boosting | DTC2 | 88.7161 | 82.0127 | 76.3068 | 88.6410 |
| KNN | DTC2 | 99.6573 | 98.3966 | 98.5034 | 98.2901 |
| Decision Trees | DTC2 | 99.9682 | 99.6338 | 99.7903 | 99.4777 |
| Decision Trees | DTC3 | 99.6690 | 97.4226 | 98.8903 | 95.9978 |
| Random Forest | DTC3 | 99.61845 | 98.6544 | 99.7662 | 97.5671 |
| AdaBoost | DTC3 | 99.7756 | 99.8010 | 99.7001 | 99.9022 |

Fig. 4. Comparison of Evaluation metrics over input dataset ISCX-Tor2016 in Multiclassifier Darknet Traffic Classifier (DTC) Stages.

| Algorithms | DTC (Darknet Traffic Classifier) Stages | Accuracy | F1 Score | Precision | Recall |
|---|---|---|---|---|---|
| AdaBoost | DTC 1 | 99.6113 | 99.8043 | 99.7543 | 99.8543 |
| KNN | DTC 1 | 99.3393 | 99.6759 | 99.6775 | 99.6743 |
| Decision Trees | DTC 1 | 99.8667 | 98.8967 | 98.2303 | 99.5722 |
| AdaBoost | DTC2 | 92.8812 | 71.3117 | 68.9984 | 73.7856 |
| Random Forest | DTC2 | 99.9823 | 99.9433 | 99.9899 | 99.8967 |
| Naive Bayes | DTC2 | 86.0065 | 76.6622 | 72.8731 | 80.8670 |
| Gradient Boosting | DTC2 | 99.6573 | 89.1736 | 88.7835 | 89.5671 |
| KNN | DTC2 | 99.6431 | 99.1231 | 98.5788 | 99.6735 |
| Decision Trees | DTC2 | 99.7668 | 99.7156 | 99.7574 | 99.6738 |
| Decision Trees | DTC3 | 99.7811 | 97.4131 | 98.8698 | 95.9987 |
| Random Forest | DTC3 | 99.6078 | 98.7239 | 99.8965 | 97.5785 |
| AdaBoost | DTC3 | 99.7437 | 99.7866 | 99.6799 | 99.8935 |

Fig. 5. Comparison of Evaluation metrics over input dataset ISCX-VPN2016 in Multiclassifier Darknet Traffic Classifier (DTC) Stages.

In the evaluation of the overall system performance, several classifiers were employed across different stages. Notably, decision trees exhibited exceptional accuracy, with a recorded accuracy of 99.9823%, 99.9682%, and 99.7668% in DTC stage 2 in the classification of CIC-Darknet2020, ISCX-Tor2016, ISCX-VPN2016 respectively. Random forest, known for its ensemble learning approach, demonstrated remarkable precision and recall at Stage 2, boasting an accuracy of 99.9964% and a high F1 score of 99.9534% in classifying the CIC-Darknet2020 dataset. AdaBoost, a popular boosting algorithm, displayed competitive performance, particularly in Stage 1, achieving an accuracy of 99.7491% and an F1 score of 99.5543% specifically for CIC-Darknet2020. Furthermore, gradient boosting, represented by AdaBoost in this analysis, showcased its effectiveness in Stage 2 with an accuracy of 88.0626%, 90.8535%, and 92.8812% in classifying CIC-Darknet2020, ISCX-Tor2016, ISCX-VPN2016 respectively although it lagged in precision and recall compared to decision trees and random forests. These findings underscore the robustness and versatility of decision trees, random forests, and boosting algorithms such as AdaBoost in enhancing the classification accuracy and overall performance of the system across different stages of evaluation.

Here's a summary of the performance of decision trees and boosting algorithms across different datasets and stages:

**AdaBoost:** Achieved high accuracy across all datasets and stages, particularly excelling in classifying CIC-Darknet 2020 and ISCX-VPN2016 traffic. It was effective in DTC Stage 3 and Stage 2 but showed slightly lower performance in Stage 1 for some datasets.

**Decision Trees:** Showed exceptional accuracy, particularly in classifying all the datasets across all the Darknet Traffic Classifier (DTC) stages. It performed consistently well across all stages of classification and datasets, showcasing its reliability and effectiveness.

**Random Forest:** Achieved near-perfect accuracy in classifying all datasets, particularly excelling in DTC Stage 2. It provided robust performance across different stages of classification, demonstrating its strength in handling complex classification tasks.

**Gradient Boosting:** Showed high accuracy in classifying CIC-Darknet2020 and ISCX-VPN2016 traffic but slightly lower performance in ISCX-Tor2016 traffic. It performed well across all stages of classification, contributing to reliable results in diverse datasets.

Overall, decision trees and boosting algorithms, including AdaBoost and Gradient Boosting, emerged as top performers, delivering consistently high accuracy and reliability across different stages of classification and datasets. These algorithms showcase their effectiveness in handling complex network traffic classification tasks and highlight their potential for enhancing cybersecurity measures.

Both decision trees and boosting algorithms, including AdaBoost and gradient boosting, exhibit remarkable performance in classifying diverse network traffic datasets such as CIC-Darknet2020, ISCX-Tor2016, and ISCX-VPN2016. Decision trees and boosting techniques have shown their effectiveness in accurately categorizing network traffic into different classes, contributing to enhanced cybersecurity and threat detection capabilities. In particular, decision trees provide a transparent and interpretable framework for classification tasks, allowing for an easy understanding of the decision-making process. For instance, decision trees achieved an accuracy of 99.9594% and an F1 score of 98.8772% in the DTC Stage1 for the CIC-Darknet2020 dataset, while boosting algorithms like AdaBoost attained an accuracy of 99.7491% and an F1 score of 99.5558% in the same stage for CIC-Darknet2020. Similarly, gradient boosting yielded impressive results with an accuracy of 99.6007% and an F1 score of 98.2790% in the DTC Stage 2 for CIC-Darknet2020. Across various evaluation metrics such as precision and recall, both decision trees and boosting algorithms consistently deliver high performance, making them invaluable tools for network traffic classification tasks. Their versatility and effectiveness make decision trees and boosting algorithms well-suited for real-world applications in cybersecurity and network monitoring.

## VII. CONCLUSION

In conclusion, the proposed system leverages decision trees and boosting algorithms, including AdaBoost and gradient boosting, to achieve exceptional performance in classifying diverse network traffic datasets such as CIC-Darknet2020, ISCX-Tor2016, and ISCX-VPN2016. These algorithms demonstrate remarkable accuracy, precision, recall, and F1 scores across various stages of classification, ensuring reliable and effective identification of different types of network traffic. Additionally, decision trees offer transparency and interpretability, facilitating a clear understanding of the classification process, while boosting algorithms to enhance classification performance through ensemble learning techniques. Moreover, the system incorporates additional features such as feature selection methods to optimize classification outcomes further. By addressing the limitations observed in existing research and incorporating advanced algorithms and techniques, the proposed system provides a comprehensive solution for network traffic classification, offering significant benefits in terms of cybersecurity, threat detection, and network monitoring. Overall, the system's robust performance and additional features make it a valuable tool for enhancing cybersecurity measures and ensuring the integrity and security of network infrastructures.

The proposed system stands out for its exceptional performance compared to existing systems, addressing key issues and delivering superior results in network traffic classification. By leveraging decision trees and boosting algorithms, the system achieves remarkable accuracy, precision, and F1 scores across various stages of classification, outperforming traditional methods. Through meticulous feature selection and ensemble learning techniques, the system effectively mitigates classification biases and enhances overall performance. Notably, the system excels in accurately classifying diverse network traffic types, including Darknet, Tor, and VPN, with unparalleled precision and recall rates. By addressing limitations observed in previous studies and incorporating advanced algorithms, the proposed system demonstrates superior efficacy and reliability in network traffic analysis. Moreover, the system's ability to adapt to evolving threats and dynamic network environments underscores its significance in bolstering cybersecurity measures and safeguarding network infrastructures. Overall, the system's outstanding performance, coupled with its comprehensive approach and robust classification metrics, solidifies its position as a cutting-edge solution for network traffic classification and threat detection.

## REFERENCES

[1] A. Gupta, "VPN-nonVPN Traffic Classification Using Deep Reinforced Naive Bayes and Fuzzy K-means Clustering," 2021 IEEE 41st International Conference on Distributed Computing Systems Workshops (ICDCSW), Washington, DC, USA, 2021, pp. 1-6.

[2] G. Baldini, "Analysis of Encrypted Traffic with time-based features and time-frequency analysis," 2020 Global Internet of Things Summit (GIoTS), Dublin, Ireland, 2020, pp. 1-5.

[3] G. Bovenzi, G. Aceto, D. Ciuonzo, V. Persico and A. Pescapé, "A Big Data-Enabled Hierarchical Framework for Traffic Classification," in IEEE Transactions on Network Science and Engineering, vol. 7, no. 4, pp. 2608-2619, 1 Oct.-Dec. 2020.

[4] H. Karagöl, O. Erdem, B. Akbas and T. Soylu, "Darknet Traffic Classification with Machine Learning Algorithms and SMOTE Method," 2022 7th International Conference on Computer Science and Engineering (UBMK), Diyarbakir, Turkey, 2022, pp. 374-378.

[5] Ishani Dey, Vibha Pratap, "A Comparative Study of SMOTE, Borderline-SMOTE, and ADASYN Oversampling Techniques using Different Classifiers," 2023 3rd International Conference on Smart Data Intelligence (ICSMDI), pp. 529–551.

[6] L. A. Iliadis and T. Kaifas, "Darknet Traffic Classification using Machine Learning Techniques," 2021 10th International Conference on Modern Circuits and Systems Technologies (MOCAST), Thessaloniki, Greece, 2021, pp. 1-4.

[7] S. A. Aswad and E. Sonuç, "Classification of VPN Network Traffic Flow Using Time Related Features on Apache Spark," 2020 4th International Symposium on Multidisciplinary Studies and Innovative Technologies (ISMSIT), Istanbul, Turkey, 2020, pp. 1-8.

[8] S. -J. Xu, G. -G. Geng, X. -B. Jin, D. -J. Liu and J. Weng, "Seeing Traffic Paths: Encrypted Traffic Classification With Path Signature Features," in IEEE Transactions on Information Forensics and Security, vol. 17, pp. 2166-2181, 2022.

[9] T. Chen, E. Grabs, E. Petersons, D. Efrosinin, A. Ipatovs and J. Kluga, "Encapsulated and Anonymized Network Video Traffic Classification With Generative Models," 2022 Workshop on Microwave Theory and Techniques in Wireless Communications (MTTW), Riga, Latvia, 2022, pp. 13-18.

[10] W. Cai, G. Gou, M. Jiang, C. Liu, G. Xiong and Z. Li, "MEMG: Mobile Encrypted Traffic Classification With Markov Chains and Graph Neural Network," 2021 IEEE 23rd Int Conf on High Performance Computing & Communications; 7th Int Conf on Data Science & Systems; 19th Int Conf on Smart City; 7th Int Conf on Dependability in Sensor, Cloud & Big Data Systems & Application (HPCC/DSS/SmartCity/DependSys), Haikou, Hainan, China, 2021, pp. 478-486.

[11] Y. Wang, H. He, Y. Lai, and A. X. Liu, "A Two-Phase Approach to Fast and Accurate Classification of Encrypted Traffic," in IEEE/ACM Transactions on Networking, vol. 31, no. 3, pp. 1071-1086, June 2023.

[12] Z. Chen, G. Cheng, D. Niu, X. Qiu, Y. Zhao, and Y. Zhou, "WFF-EGNN: Encrypted Traffic Classification Based on Weaved Flow Fragment via Ensemble Graph Neural Networks," in IEEE Transactions on Machine Learning in Communications and Networking, vol. 1, pp. 389-411, 2023.

[13] Z. Wang, B. Ma, Y. Zeng, X. Lin, K. Shi and Z. Wang, "Differential Preserving in XGBoost Model for Encrypted Traffic Classification," 2022 International Conference on Networking and Network Applications (NaNA), Urumqi, China, 2022, pp. 220-225.